\documentclass[conference]{IEEEtran}
\makeatletter
\newcommand{\newlineauthors}{%
  \end{@IEEEauthorhalign}\hfill\mbox{}\par
  \mbox{}\hfill\begin{@IEEEauthorhalign}
}
\makeatother
\IEEEoverridecommandlockouts

\usepackage{cite}
\usepackage{amsmath,amssymb,amsfonts}
\usepackage{algorithmic}
\usepackage{graphicx}
\usepackage{textcomp}
\usepackage[table]{xcolor}
\usepackage{microtype}
\usepackage{tabularx}
\usepackage{multicol}
\usepackage{multirow}
\usepackage{array}
\usepackage{comment}
\usepackage{booktabs}
\usepackage{stfloats} 
\usepackage{cleveref}
\usepackage{siunitx}
\usepackage{threeparttable}
\newcolumntype{C}{>{\centering\arraybackslash}X}

\usepackage[prependcaption,textsize=scriptsize]{todonotes}
\definecolor{mycolor}{HTML}{FF6600}

\def\BibTeX{{\rm B\kern-.05em{\sc i\kern-.025em b}\kern-.08em
    T\kern-.1667em\lower.7ex\hbox{E}\kern-.125emX}}
\begin{document}

\title{Self-supervised Speech Comparison for\\ L2 Phone, Rhythm, and Intonation Scoring} 

\author{\IEEEauthorblockN{
    Stephen McIntosh\IEEEauthorrefmark{2}$^*$\thanks{$^*$These authors contributed equally.},
    Reuben Smit\IEEEauthorrefmark{3}$^*$,
    Daisuke Saito\IEEEauthorrefmark{2},
    Nobuaki Minematsu\IEEEauthorrefmark{2} and
    Herman Kamper\IEEEauthorrefmark{3}
}
\IEEEauthorblockA{\IEEEauthorrefmark{2}The University of Tokyo, Japan \,\, \IEEEauthorrefmark{3}Stellenbosch University, South Africa\\
\{smcintosh, dsk\_saito, mine\}@gavo.t.u-tokyo.ac.jp \,\, \{28583272, kamperh\}@sun.ac.za}
}


\maketitle

\begin{abstract}
L2 speech assessment has traditionally focused on phonetic assessment, leaving the scoring of suprasegmental
features such as rhythm and intonation underexplored. Moreover, assessment methods often require training
with labeled L2 speech data, making them difficult to apply in low-resource settings. We investigate whether
DTW over self-supervised WavLM representations can provide a text-free framework for assessing
phonetic accuracy, rhythm, and intonation in English and Japanese L2 speech. Results show that
a basic DTW-based approach that compares learner speech to native templates exceeds
human agreement on holistic and sentence-level phonetic scoring.
For rhythm, we introduce methods that measure the degree of warping in the
DTW alignment path; our best method approaches human-level performance.
For intonation, we combine DTW distance over prosodic residuals with pitch and intensity features, but performance remains more modest on some tasks. Our results point to self-supervised representations as a promising, text-free basis for multi-aspect pronunciation assessment. 
\end{abstract}


\section{Introduction}

Most work in automatic second-language (L2) speech evaluation focuses on segmental features (i.e., phones), disregarding suprasegmental (prosodic) features such as rhythm and intonation~\cite{lounis2024mispronunciation}. 
However, suprasegmental aspects of speech play a vital role in how listeners process and understand spoken language~\cite{cutler1997prosody, dahan2015prosody, bae2015effects}. 
Skilled expression of these speech aspects is known to improve comprehensibility and reduce accentedness of L2 speech~\cite{tajima1997effects, derwing1998evidence,van2021interplay}.

\begin{figure}[!b]
  \centering
  \includegraphics[width=0.9\linewidth]{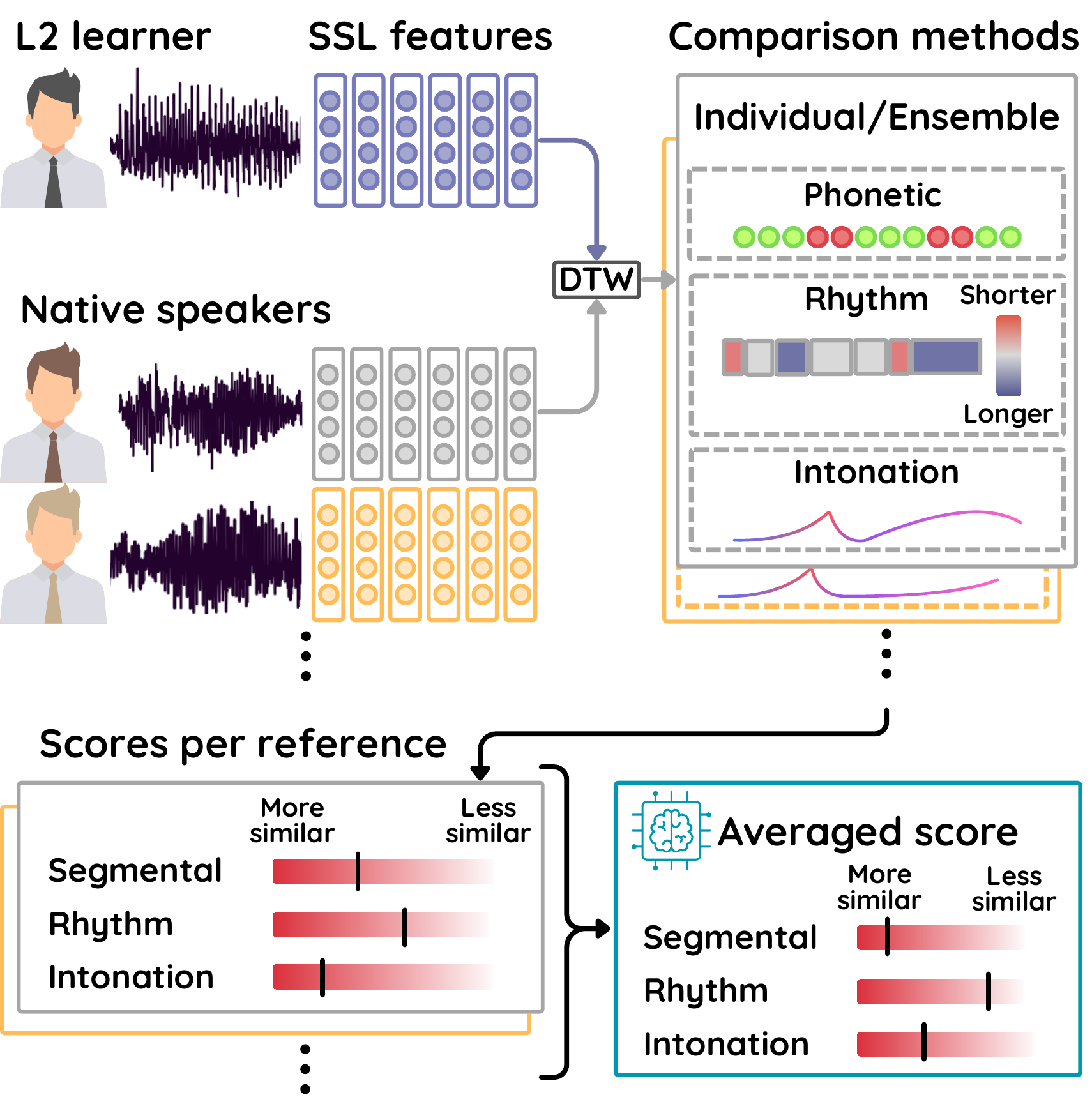}
  \caption{Our L2 pronunciation assessment approach. Self-supervised speech representations from second-language learners are compared with native references to identify differences in phones, rhythm, and intonation. 
  The raw scores are then converted to ordinal scores.
  }
  \label{fig:summary}
\end{figure}

In this work, we predict phonetic and prosody scores of L2 utterances by comparing their self-supervised speech representations to those of native speech templates.
Our methods only use a few native templates of the target content and are therefore viable for L2 assessment in low-resource languages where gathering scored L2 speech is challenging.
Our approach is summarized in~\Cref{fig:summary}: 
We align self-supervised speech representations of native templates with L2 learner speech using dynamic time warping~(DTW).
We use this alignment in various comparison methods, each of which targets 
phonetics, rhythm, or intonation.
We then average the raw results of these methods over all native references into one representative comparison score.
To evaluate each method, we calculate the correlations of these aggregate scores with human ratings, and compare with pairwise human agreement.

In experiments on English and Japanese, we demonstrate that self-supervised speech representations combined with DTW provide useful information for automated pronunciation scoring. For phonetic assessment, they perform well and sometimes exceed expert inter-rater agreement.
For rhythm, we show that the warping path itself is a rich source of information: our two novel rhythm metrics approach human inter-rater agreement, while significantly outperforming the conventional baseline.
We introduce $k$-means residuals of SSL features as a prosodic signal for intonation scoring, and show that they outperform $F_0$ and intensity features, although overall performance is modest when compared to human experts. 

Our primary contribution is the introduction of novel DTW-based approaches for assessing suprasegmental speech aspects in L2 speech.
To our knowledge, we also provide the first comparison of template-based assessment methods to inter-human agreement for both segmentals and suprasegmentals.
Upon acceptance, we will release our methods and evaluation~code. 

\section{Related Work}\label{sec:related-work}

Automatic pronunciation assessment has been a core component of computer-assisted pronunciation systems for several decades.
It has broadly fallen into two paradigms: acoustic-model-based approaches, and template-based approaches. 

Of the \textbf{acoustic-model-based approaches}, the most well-known is the goodness of pronunciation (GOP) score~\cite{witt2000GOP}. This is a likelihood-based approach that compares the produced phones with what is expected,
flagging mispronunciations when the likelihood falls below a threshold. 
Building on this idea, subsequent work focused on improving goodness of pronunciation-related metrics by using improved architectures and deep learning techniques~\cite{strik2009comparing, hu2013DNNHMM, shi2020context, yan2020end, gong2022GOPT}.
While effective, this requires large amounts of labeled data to train acoustic models, a phoneme inventory for the language, and a canonical transcription for each assessed utterance.
This is unfeasible in languages with sparse resources and a lack of functional speech recognition. 

\textbf{Template-based approaches} compare the learner's speech directly against a parallel native reference utterance by extracting frame-level representations and aligning them, often using dynamic time warping (DTW).
This paradigm requires no phoneme inventory, canonical transcription, or additional model training, making it applicable to low-resource settings.
Early work used
mel-frequency cepstral coefficients (MFCCs) as input
representations~\cite{lee2012comparison, martijn2020mfccbaseline}. 
Subsequent frameworks extracted geometric and duration features from the DTW path and distance matrices to perform pronunciation assessment~\cite{lee2013pronunciation}. Phonetic posteriorgrams (PPGs)~\cite{lee2012comparison, lee2013mispronunciation} occupy an interesting middle ground between the two paradigms: an acoustic model generates frame-level posterior distributions over a fixed phoneme inventory, which are then compared using DTW~\cite{PPGDTW}. 
This preserves template-based comparisons while introducing structured phonetic representations, at the cost of reintroducing a dependency on a target-language acoustic model.

Recent template-based approaches have moved toward \textbf{self-supervised learning} (SSL) models such as wav2vec 2.0~\cite{wav2vec2} and HuBERT~\cite{hubert}, which are pre-trained on large quantities of unlabeled speech and learn rich, hierarchically organized representations without annotations.
Bartelds et al.~\cite{bartelds2022neural} demonstrated that SSL features combined with DTW outperform MFCC-based comparison for measuring phonetic distances between speech varieties. 
SSL representations also encode suprasegmental features such as lexical stress, pitch accent, and tone~\cite{de2024layer, prosodicABX}. 
In particular, Sun et al.~\cite{prosodicABX} demonstrate that SSL-based template comparison is sensitive to suprasegmental contrasts.
These findings suggest that SSL features are not only fit for segmental scoring, but are capable of capturing the full range of pronunciation\,---\,segmental and suprasegmental alike. 
With this in mind, we explore template-based approaches for assessing L2 speech along segmental and suprasegmental axes.

\section{Speech Scoring Methods}\label{sec:methods}
We propose a range of methods for scoring phonetics and prosody. Where phonetics (the segmentals) relate to \textit{what} you say, prosody (the suprasegmentals) pertains to \textit{how} you say it. We further split prosody into intonation and rhythm. 

As seen in \Cref{fig:summary}, our methods compare learner speech to a native template using DTW alignment of WavLM-Large~\cite{wavlm} representations. We define the distance between two frames by their cosine distance. We apply DTW, which temporally aligns the learner and native-speaker utterances to minimize the sum of the distances between aligned frames. This alignment is computed as a path between the learner utterance on the $x$-axis and template utterance on the $y$-axis (see the left of~\Cref{fig:warp-angle}).

Each method uses this alignment differently: phone scoring uses the DTW distance directly, rhythm scoring uses the path, and intonation scoring uses alternative distance metrics 
over the warp path. Since our methods are reference-based, we compute a similarity score for every native template available in the respective task subset, and average all those scores.

\subsection{Phone scoring}
Phones refer to the smallest discrete segments of sound in speech, such as consonants and vowels. Self-supervised speech representations have been shown to have significant phonetic information~\cite{wav2vec2, hubert}, so we simply use the normalized DTW distance over representations from the final layer in WavLM-Large to compute a phone score, as in previous work (see \Cref{sec:related-work}). Concretely, we take the final aggregated DTW alignment cost, i.e., the cumulative sum of the cosine distances along the optimal path, normalized by the path length. 

\textit{Baselines}: To quantify the contribution of self-supervised features, we also compare with phone scores computed using phonetic posteriorgram (PPG) features~\cite{churchwell2024PPG} with Jensen-Shannon distance.

\subsection{Rhythm scoring}\label{sec:method-rhythm}

Rhythm refers to the perceived macro-temporal pattern of language, or in other words, the recurrence of linguistic units of speech at regular intervals. While global speech rate is certainly perceived in language, we focus on identifying rhythmic errors by identifying local deviations in speech rate from the typical rhythm exemplified by the native template. We propose two methods: tempo irregularity, which measures small deviations in
the warping path, and interval distortion, which compares the durations of vowel or consonant clusters between the two tracks.

\begin{figure}[!t]
  \centering
  \includegraphics[width=\linewidth]{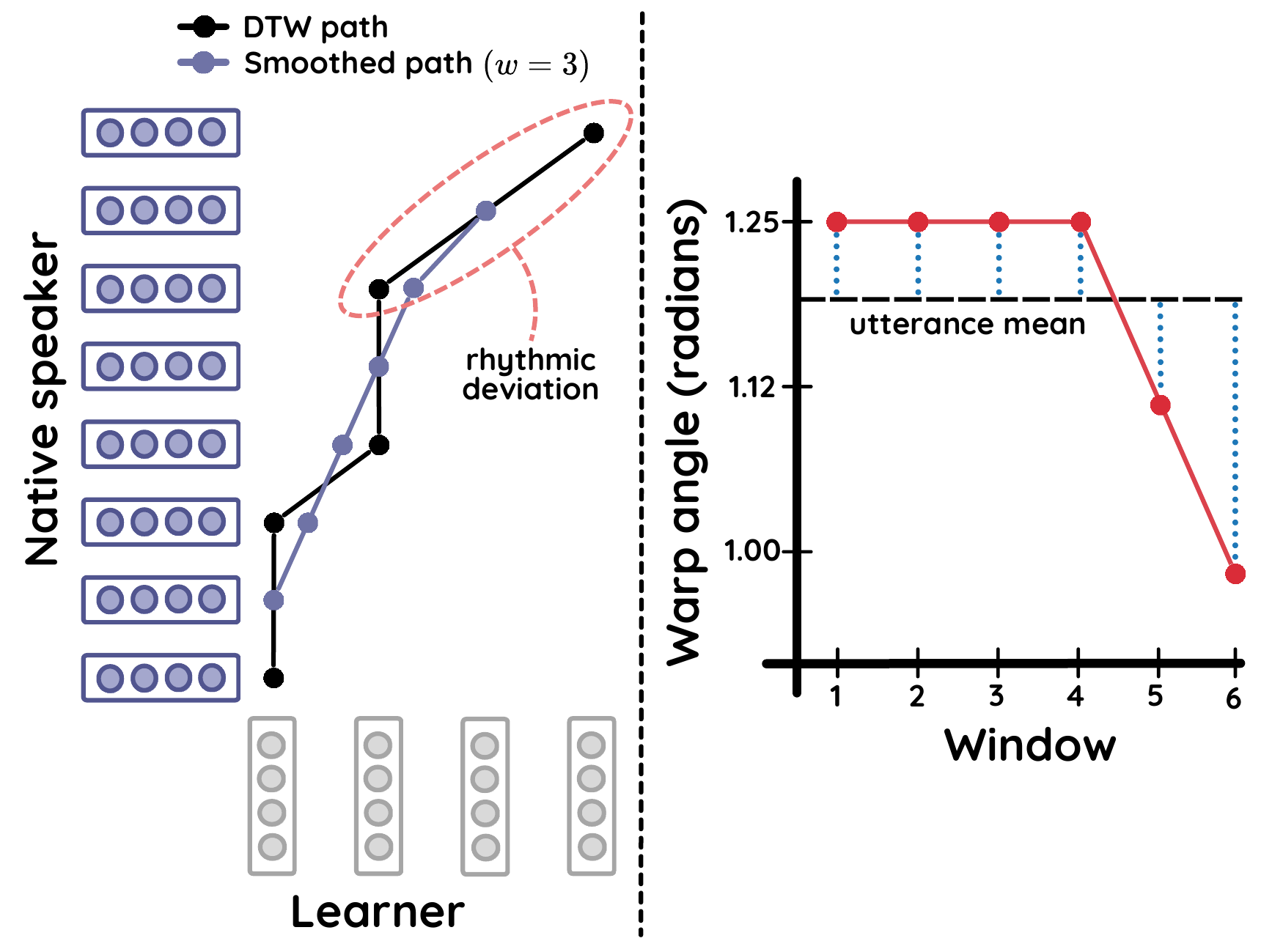}
  \caption{Our method for determining the warp angle of a DTW path. To smooth quantization noise in the warping path (left), we calculate the average of overlapping frame windows, resulting in a smoothed warping path. We then calculate the angle in radians over 3-frame windows (right). Our tempo irregularity is the average distance of the angles from the utterance mean.
  }
  \label{fig:warp-angle}
\end{figure}

\textbf{Tempo irregularity} uses only the warp path as input. Horizontal steps in this path represent time dilation (the learner is locally slower than the template), and vertical steps represent time compression (the learner is locally faster than the template). 
We define the \textit{warp angle} as the instantaneous pacing ratio between the two utterances. 
To capture rhythmic irregularity, we first approximate the warp angle at each frame and then calculate the dispersion of these angles about their mean.

\begin{figure}[!t]
  \centering
  \includegraphics[width=\linewidth]{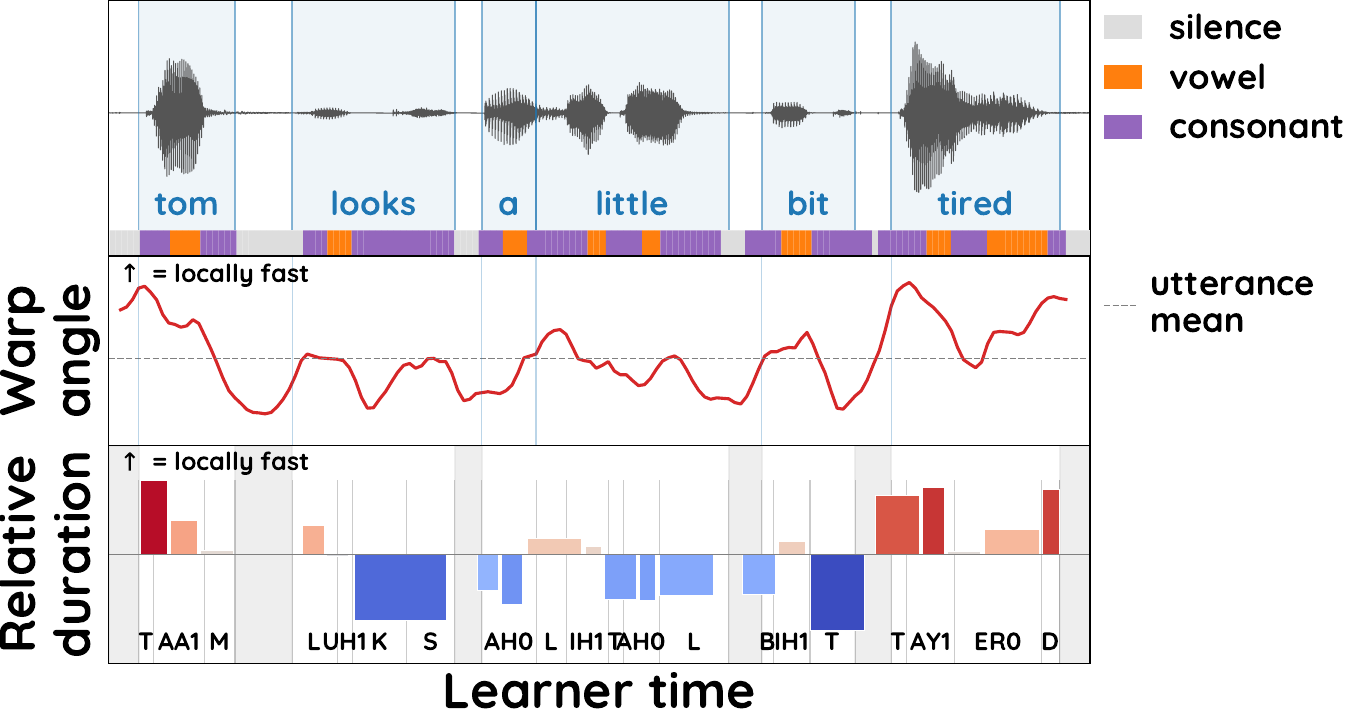}
  \caption{An example of our two rhythm assessment methods. For tempo irregularity, we smooth the warping path over 5-frame windows and calculate the deviation of the angle of that path from its mean (red line). For the interval deviation method, frames of speech are classified as silence, vowels, or consonants. We group contiguous phones of each class, and calculate the difference in log-duration between each native and learner interval.
  }
  \label{fig:rhythm}
\end{figure}

The simplest method to approximate the warp angle is to use a short overlapping window. 
To approximate the warp angle without quantization noise from discrete DTW steps, we first map each template frame to its average corresponding learner frame, producing a sequence of the same length as the template. Then we smooth this using a $5$-frame (100 ms) moving average (example shown for a 3-frame window in blue on the left of \Cref{fig:warp-angle}), approximate the derivative at each point, and calculate the corresponding angles.
The result is
a sequence of angles $a = \{a_3, \ldots, a_{T - 2} \mid a_i \in \left[0, \frac \pi 2\right)\}$, shown on the right of \Cref{fig:warp-angle} and again in the middle tier of \Cref{fig:rhythm}.

We calculate the absolute difference of each angle from the utterance mean. The average of those values represents the rhythmic deviation of the L2 speaker from the native reference.

\textbf{Interval distortion} is inspired by work in linguistics on speech rhythm.
Speech rhythm was once thought to be determined by the duration between stresses or syllables, depending on whether a language is stress-timed or syllable-timed~\cite{pike1945intonation, abercrombie1967rhythm}. 
However, it has been shown that inter-syllable and inter-stress durations are not always consistent measures of rhythm~\cite{dauer1983stress, ramus1999rhythm}. In recent linguistic analyses of rhythm, 
speech is divided into vocalic and non-vocalic intervals, and speech rhythm is characterized based on the durations of those intervals~\cite{ramus2002acoustic}. We do the same with learner and native speech, using the SSL representations to determine which parts of the speech are vowels, consonants, or silence. 

We train two logistic regression models on TIMIT~\cite{TIMIT} to classify frames: one for silence and one for vowels. At inference time, the silence classifier first filters out non-speech regions, and the vowel classifier then distinguishes vocalic from consonantal segments among the remaining frames, as seen in \Cref{fig:rhythm}.
For each such interval in the template, the warping path determines how many learner frames it maps onto. The log-ratio of learner to native duration is computed per interval, as can be seen in the bottom tier of \Cref{fig:rhythm}, where positive segments (red) are fast relative to the template, and negative segments (blue) are slow.
As with tempo irregularity, we calculate the absolute difference of these relative durations from their mean, and take their average over the utterance. This final metric measures how well the learner matched native speakers' local rhythm.

It is often assumed~\cite{grabe2002rhythm} that only vowel segments expand or contract with speech rate, but interval distortion performed better in preliminary experiments when using both consonants and vowels, consistent with evidence that speakers compress or expand entire syllable blocks to maintain timing~\cite{tajima2004speech}.

\textit{Baselines}: We report the absolute \textbf{duration ratio} (${R_{\text{dur}}}$) between the learner and native utterances, which approximates the difference in speech rate between the two speakers. This helps determine the influence of global speech rate.
Using the vocalic interval durations above, we also calculate the normalized pairwise variability index ($\text{nPVI}_\text{V}$)~\cite{grabe2002rhythm}. $\textbf{nPVI}_\textbf{V}$ summarizes variation in duration between successive vocalic intervals.

\subsection{Intonation scoring}
Intonation relates to the changes in the manner of speech (mostly pitch) over time.
Informed by previous work showing that prosodic and speaker information are contained in the $k$-means residuals of self-supervised features~\cite{kmeansresiduals, aihara2026exploring}, we hypothesize that calculating the distance using the z-normalized $k$-means residuals over the warp path could provide a good intonation metric.

\begin{figure}[!t]
  \centering
  \includegraphics[width=0.9\linewidth]{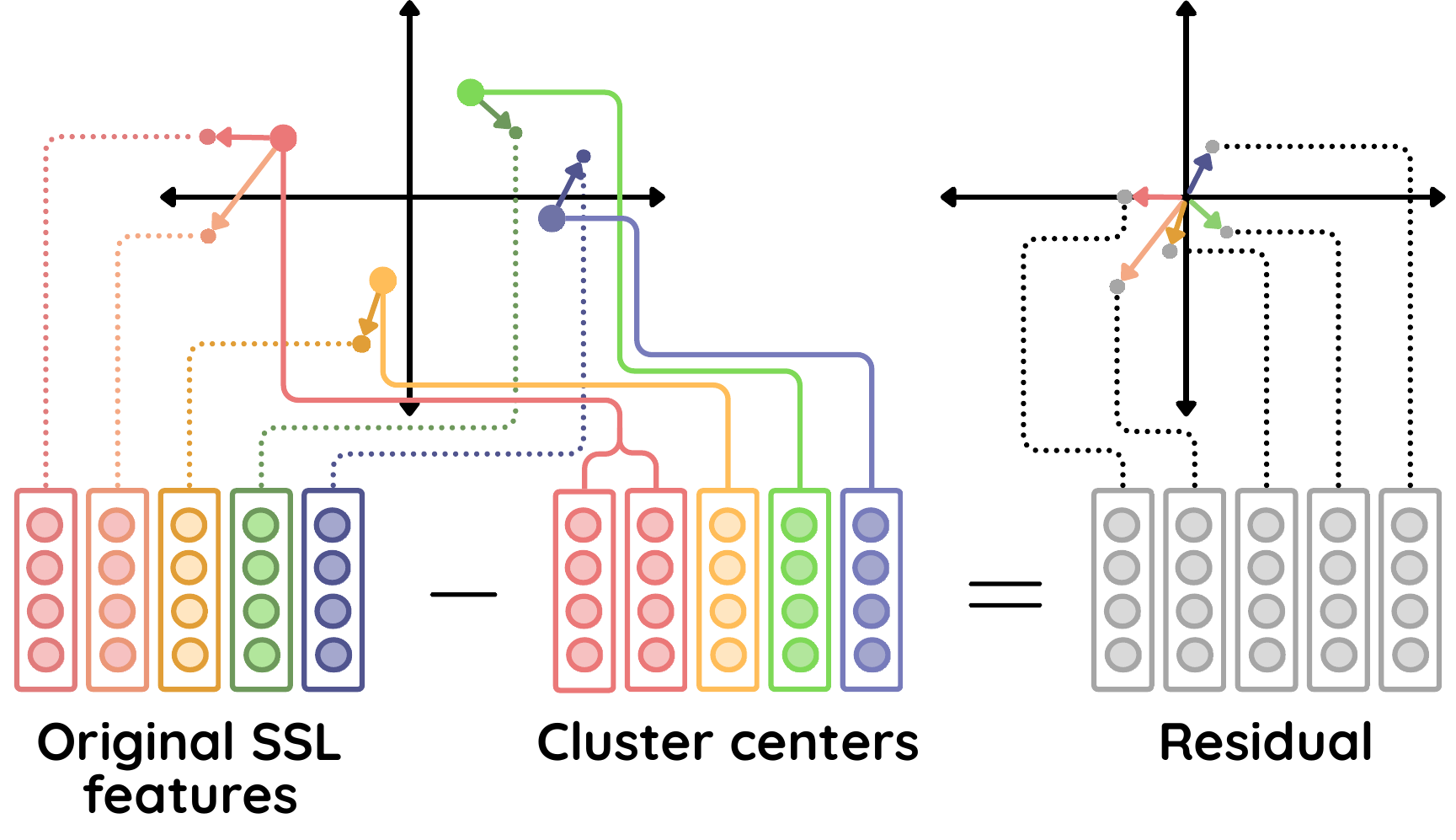}
  \caption{Our method for extracting prosodic information from SSL representations. The $k$-means cluster centers corresponding to each feature are subtracted from the original features, resulting in residual vectors. 
  }
  \label{fig:residuals}
\end{figure}

We train a $k$-means model with 200 cluster centers on a 10-hour subset of LibriSpeech train-clean-100~\cite{librispeech} for the 6th and 24th layers of WavLM-Large, following previous work~\cite{knnvc, linearvc, aihara2026exploring}.
We use these to compute residual features, as illustrated in \Cref{fig:residuals}: we replace each frame with its closest cluster center, representing it by its canonical embedding. Subtracting this from the original representations results in residuals that ideally contain a mixture of speaker-specific and prosodic information.
The mean residual over the utterance contains speaker-specific information, while the standard deviation ($\sigma_\text{res}$) is an approximation of the dynamic range over the entire utterance~\cite{kmeansresiduals}.
A very expressive speaker will have a larger standard deviation than a more somber one; to
suppress speaker-specific information while normalizing for differing dynamic ranges between speakers, we subtract the mean and divide by the standard deviation, resulting in  \textbf{z-normalized residuals} that ideally contain mostly local prosodic information.
We report the DTW distance along the warp path between these prosodic residuals ($\text{DTW}_\text{norm. res}$), and the standard deviation of the residuals $\sigma_\text{res}$.

\textit{Baselines}: We calculate two features: $F_0$ and $F_0$--intensity, using torchcrepe~\cite{morrison2020torchcrepe}. $F_0$ is log-transformed and mean-centered. Since it improved performance in early experiments, unvoiced frames are filled with interpolated $F_0$ values. For each of these features, we calculate utterance scores using the Euclidean DTW distance.

\begin{table*}[!b]
\renewcommand{\arraystretch}{1.0}
\centering
\begin{threeparttable}
\caption{Summary of the rating tasks in the ERJ and JRF read-speech
corpora. 
All tasks are rated on a 5-point scale.
\emph{Materials} is the symbolic guidance shown to the speaker during recording;
\emph{Texts} is the number of distinct rated text prompts;
\emph{L1/text} and \emph{L2/text} give the number of distinct rated speakers per
prompt.}
\label{tab:datasets}
\small
\begin{tabularx}{\textwidth}{@{}lXXccc@{}}
\toprule
Task & Materials & What the rater judged & Texts & L1/text & L2/text \\
\midrule
\rowcolor{gray!15}[0pt][0pt]
\multicolumn{6}{@{}l}{\textbf{ERJ}: English Speech Database Read by Japanese Students} \\
\addlinespace[1pt]
Phones (sentence)     & Text (phonemes on practice sheet) & \emph{Phonetic content} & 625 & 10--11 & 1--6 \\
Rhythm (sentence)    & Text + sentence stress & Does the \emph{rhythm} match the materials? & 120 & 10--11 & 6--10 \\
Intonation (sentence) & Text + intonation arrows & Does the \emph{intonation} match the materials? & 60 & 9 & 14--16 \\
Phones (word)         & Text + phonemes & \emph{Phonetic content} & 300 & 8--9 & 11--14 \\
Accent (word)         & Text + phonemes + lexical stress & \emph{Lexical stress} & 77 & 8 & 24--26 \\
\midrule
\rowcolor{gray!15}[0pt][0pt]
\multicolumn{6}{@{}l}{\textbf{JRF}: Japanese Speech Database Read by Foreign Students} \\
\addlinespace[1pt]
Overall score (sentence)              & Spaced text + kana$^\ddagger$ & \textit{Pronunciation} (holistic scoring) & 30 & 20--21 &
19--26 \\
Intonation (sentence)$^\dagger$ & Dialogue text + coaching during practice & 1--3 text-specific \textit{intonation} criteria 
(e.g., prominence, question-rise, fillers) & 8 & 40--41 & 141 \\
Difficult sounds (word)         & Text + kana$^\ddagger$ & Text-specific \emph{segmental} or \emph{timing} feature
(e.g., gemination, long vowel) & 10 & 41 & 140--141 \\
\bottomrule
\end{tabularx}

\begin{tablenotes}
        \item[$\dagger$] A subset of the JRF C-set prosody ratings including only \emph{intonation}-related criteria (phrasing/boundary criteria excluded). Criteria were averaged for sentences with more than one to create a single score for each utterance.
        \item[$\ddagger$] A Japanese script that completely determines the phoneme sequence.
    \end{tablenotes}
\end{threeparttable}
\end{table*}

\subsection{Overall scoring approach}
For each method described above, we calculate an average score over all available native templates. 
The resulting \textit{raw scores} are a distance measure between the learner's speech and all the available native templates for phones, rhythm, or intonation. 

As rater scores might depend on information captured to different degrees by different methods, we additionally calculate ensemble scores. These 
are computed by combining the raw scores of a few methods using a linear regression model to produce a single representative distance-like score that will maximize the evaluation metrics within each cross-validation fold.

\section{Evaluation}
\subsection{Datasets and tasks}

We use two datasets for evaluation: English read by Japanese students (ERJ)~\cite{ERJ} and Japanese read by foreign students (JRF)~\cite{JRF}. These have similar designs: they are parallel corpora of L1 and L2 read speech, with subsets targeting different aspects of speech (like phones or prosody). L2 speech is scored in each of these subsets by expert raters. No inter-rater calibration was performed.

\Cref{tab:datasets} summarizes the subsets used in our evaluation, including the task, materials provided during recording, rating criteria, and the number of distinct texts and corresponding utterance counts. While no tasks are parallel, these represent a rich set of phonetic and suprasegmental features.

We evaluate each method with speaker-disjoint 5-fold cross-validation: utterances are scored by a model fit only on the folds that contain none of its speaker's utterances. This uses all data for evaluation and avoids dependence on a single random~split.

\subsection{Evaluation metrics}

We calculate \textbf{Pearson correlation coefficient} ($r$) and \textbf{Spearman's rank correlation} ($\rho$), which respectively measure the linear and monotonic agreement between the raw metric values and rater scores. These reflect the quality of the underlying representations for each task.

To evaluate how well raw scores can be used to predict raters' ordinal scores, we train linear regression models to convert the raw scores of individual and ensemble methods to ordinal scores by tuning within each cross-validation fold.
These discretized ordinal classes are evaluated using \textbf{quadratic weighted kappa} (QWK), which measures agreement between the discretized predictions and human scores, penalizing disagreement in proportion to their squared distance. We also report \textbf{mean absolute error} (MAE). Since these absolute-agreement metrics are confounded by per-rater differences in offset and scale, we focus on Pearson's $r$, which is invariant to affine differences.

Each utterance receives one model prediction, from the cross-validation fold in which its speaker was held out. We pool these out-of-fold predictions and compare them to the scores of each human rater separately, computing all four metrics per rater and then averaging across raters. To contextualize these numbers, we compute a human benchmark the same way, but comparing raters to each other instead of to the model: for every pair of raters, we compute the four metrics between their scores and then average over all pairs. We assess statistical significance with a paired speaker-cluster bootstrap on $r$ using 10{,}000 resamples.

To understand which features drive ensemble performance, we also apply a Shapley-based attribution framework to each ensemble model~\cite{lundberg2017SHAP}. Shapley values decompose the model's explained variance additively across input features, providing a principled measure of each feature's marginal contribution.

\section{Results}

\begin{figure}[!b]
    \centering
    \includegraphics[width=\linewidth]{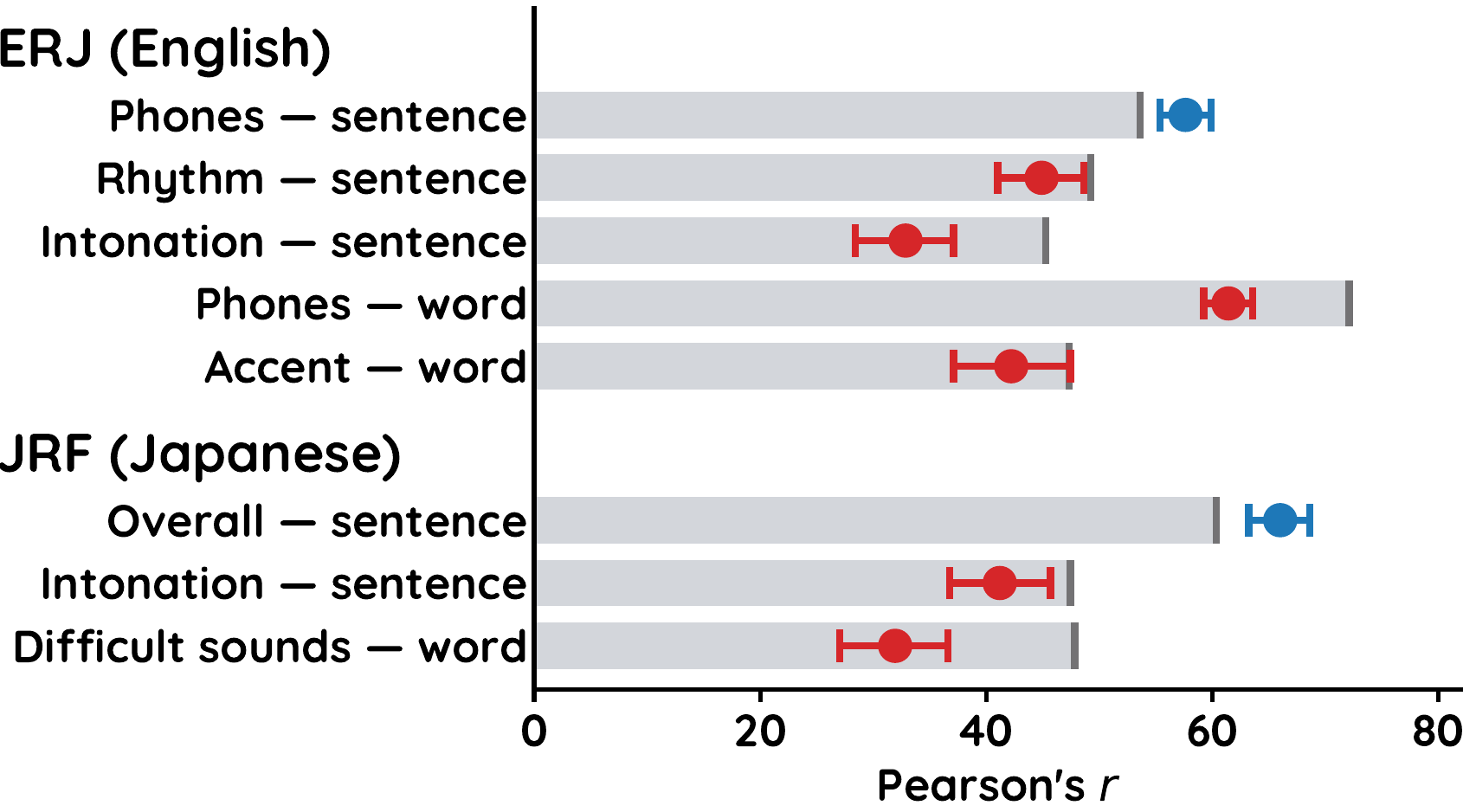}
    \caption{Best model vs. the human benchmark on each task.
     Dots are \textcolor[HTML]{1f77b4}{\textbf{blue}} when the model exceeds human agreement and \textcolor[HTML]{d62728}{\textbf{red}} otherwise. 
    Whiskers show the $95\%$ confidence interval of $\text{model} - \text{human}$; a whisker clear of the benchmark \textcolor[HTML]{737274}{\textbf{tick}} indicates a statistically significant difference.
    }
    \label{fig:pcc-vs-human}
\end{figure}

We first look at the overall best results before turning to a detailed analysis. For all assessed tasks, our DTW-based methods comfortably outperform the baselines and largely approach human agreement.

Evaluation results for the best method on each task are summarized in \Cref{fig:pcc-vs-human}. SSL-based methods perform well overall, most strikingly for phonetic scoring of sentences, where DTW over SSL features significantly exceeds human inter-rater agreement. 
Phonetic scoring of isolated words is weaker, consistent with the reduced averaging and alignment constraints that short sequences impose on DTW-based methods.
For tasks with a rhythmic component (i.e., English sentence rhythm and English word accent), our methods fall only slightly below human agreement, though the gap for word accent is not statistically significant, indicating that the rhythm scoring approach transfers well to both tasks. 
Intonation is the most challenging aspect across both languages: our best methods fall short of human agreement, and the acoustic baselines perform significantly worse. This suggests that the perceptual dimensions underlying expert intonation judgments are not well captured by either pitch-based features or SSL prosodic residuals in their current form, and that more targeted prosodic representations will be needed.

Template ablation reveals a consistent picture: for the best-performing method in each task, $\leq$5 native templates recover $\geq$95\% of the full-template agreement. Only 2 are required for English phonetic scoring and Japanese intonation. The sole exception is the JRF difficult-sounds task, which requires 8. However, the relative rankings of scoring methods remain stable regardless of the number of templates used.

We now examine each task in turn, with full per-method breakdowns in Tables~\ref{tbl:results-erj} and \ref{tbl:results-jrf}.

\subsection{ERJ (English)}
\begin{table}[!b]
\renewcommand{\arraystretch}{1.0}
\setlength{\tabcolsep}{2.5pt}
\begin{threeparttable}
\caption{Evaluation results for tasks in the ERJ~\cite{ERJ} dataset (English)}
\label{tbl:results-erj}
\begin{tabularx}{\linewidth}{@{}X S[table-format=-2.1] S[table-format=-2.1] S[table-format=-2.1] S[table-format=1.2]@{}}
\toprule
\textbf{Task}/Method & {$r$ ($\uparrow$)} & {$\rho$ ($\uparrow$)} & {QWK ($\uparrow$)} & {MAE ($\downarrow$)}\\
\midrule
\textbf{Phones (sentence)}\\
\hspace{2mm}\textit{Human benchmark} & 53.6 & 51.5 & 49.1 & 0.75 \\
\vspace{-0.08in}\\
\hspace{2mm}$\text{DTW}_{\text{SSL}}$$^{\dagger}$ & \bfseries 57.6 & 55.8 & 44.7 & 0.64 \\
\hspace{2mm}$\text{DTW}_\text{PPG}$ & 36.9 & 36.5 & 18.3 & 0.72 \\
\vspace{-0.07in}\\

\textbf{Rhythm (sentence)}\\
\hspace{2mm}\textit{Human benchmark} & 49.3 & 48.2 & 45.9 & 0.81 \\
\vspace{-0.08in}\\
\hspace{2mm}$R_\text{dur}$ + Interval distortion$^{*}$ & \bfseries 44.9 & 45.8 & 33.8 & 0.73 \\
\hspace{2mm}Interval distortion & \bfseries 43.6 & 44.9 & 31.3 & 0.74 \\
\hspace{2mm}$R_\text{dur}$ + Tempo irregularity$^{*}$ & \bfseries 43.5 & 44.2 & 32.4 & 0.73 \\
\hspace{2mm}Tempo irregularity & 37.3 & 39.0 & 25.9 & 0.76 \\
\hspace{2mm}$R_\text{dur}$ (baseline) & 31.7 & 30.3 & 21.8 & 0.78 \\
\hspace{2mm}$\text{nPVI}_\text{V}$ (baseline) & 16.5 & 16.5 & 10.2 & 0.84 \\
\vspace{-0.07in}\\

\textbf{Intonation (sentence)}\\
\hspace{2mm}\textit{Human benchmark} & 45.3 & 44.8 & 38.6 & 0.98 \\
\vspace{-0.08in}\\
\hspace{2mm}$\text{DTW}^{\text{L6}}_\text{norm. res}$ + $\sigma^{\text{L6}}_{\text{res}}$ + $F_0$--intensity$^{*}$ & \bfseries 32.9 & 32.5 & 15.5 & 0.87 \\
\hspace{2mm}$F_0$--intensity (baseline) & 21.9 & 21.0 & 2.5 & 0.89 \\
\hspace{2mm}$\sigma_{\text{res}}$ & 18.7 & 18.6 & 3.9 & 0.89 \\
\hspace{2mm}$F_0$ (baseline) & 16.9 & 18.3 & 0.9 & 0.90 \\
\vspace{-0.07in}\\

\textbf{Phones (word)}\\
\hspace{2mm}\textit{Human benchmark} & 72.1 & 72.0 & 71.3 & 0.60 \\
\vspace{-0.08in}\\
\hspace{2mm}$\text{DTW}_{\text{SSL}}$ & \bfseries 61.4 & 64.0 & 51.7 & 0.70 \\
\hspace{2mm}$\text{DTW}_\text{PPG}$ & 37.6 & 38.7 & 24.4 & 0.85 \\
\vspace{-0.07in}\\

\textbf{Accent (word)}\\
\hspace{2mm}\textit{Human benchmark} & 47.3 & 41.5 & 35.0 & 0.95 \\
\vspace{-0.08in}\\
\hspace{2mm}$R_\text{dur}$ + TI + $\sigma_{\text{res}}$ + $\text{DTW}_\text{norm. res}$$^{*}$ & \bfseries 42.2 & 41.4 & 23.9 & 0.76 \\
\hspace{2mm}$R_\text{dur}$ (baseline) & 20.2 & 21.6 & 4.4 & 0.80 \\
\hspace{2mm}$F_0$ (baseline) & 0.3 & -3.9 & 0.0 & 0.81 \\
\hspace{2mm}$F_0$--intensity (baseline) & -2.9 & -3.4 & 0.0 & 0.81 \\
\bottomrule
\end{tabularx}
\begin{tablenotes}[flushleft]
\footnotesize
\item $^{*}$Ensemble method. \textbf{Bold}: best or statistically equivalent (speaker-cluster bootstrap, $p>0.05$). $^{\dagger}$Significantly exceeds the human benchmark ($p<0.05$).
\end{tablenotes}
\end{threeparttable}
\end{table}

\textbf{Phones:} For phonetic scoring of sentences, standard DTW distance on SSL features ($r=57.6$) alone significantly exceeds human agreement ($r=53.6$). However, the results are worse for the phonetic scoring of words, which is understandable, as SSL models rely on context in a longer utterance to improve the quality of representations. 

\textbf{Sentence rhythm:}
The best-performing rhythm model combines $R_{\text{dur}}$ and interval distortion ($r = 44.9$, the highest among suprasegmental tasks), though interval distortion alone ($r=43.6$) is statistically indistinguishable from it. These are close because interval distortion already encodes speech rate (Pearson correlation is $0.44$ with $R_{\text{dur}}$), so the inclusion of duration does not contribute much (28.8\% of the ensemble's explained variance). In contrast, tempo irregularity is nearly rate-invariant ($0.24$ correlation with $R_{\text{dur}}$), so it benefits more from ensembling with duration (which is responsible for 40.3\% of the explained variance). All of the best methods incorporate both global speaking rate and local timing information, consistent with findings that more proficient speakers speak more quickly and have more appropriate local rhythm~\cite{galaczi2017assessing}.

\textbf{Intonation:}
Our best-performing intonation method achieves $r=32.9$ (human benchmark $r=45.3$), among the lowest of all evaluated tasks. Shapley attribution 
indicates that $\text{DTW}_\text{norm. res}^{\text{L6}}$ is the dominant contributor (41.0\%), followed by $F_0$–intensity features (31.2\%) and $\sigma_\text{res}^{\text{L6}}$ (27.8\%). Despite achieving a lower MAE than the human benchmark (0.87 vs. 0.98), the correlation gap remains substantial, suggesting that neither the acoustic baseline nor the SSL prosody features capture the perceptual dimensions of intonation that human raters rely on.

\textbf{Word accent:}
The best-performing word accent model combines multiple rhythm and intonation methods, achieving $r = 42.2$, which is not significantly different from the human scores of $r=47.3$ (\Cref{fig:pcc-vs-human}). 
Shapley decomposition shows $\text{DTW}_\text{norm. res}$ as the largest contributor (39.3\%), followed by tempo irregularity (denoted TI, 27.7\%), with measures targeting rhythm and intonation each accounting for half of the explained variance.
This is consistent with lexical accent in English, where a combination of vowel elongation (affecting rhythm) and pitch inflection (affecting intonation) is used to stress syllables.

\subsection{JRF (Japanese)}
\textbf{Overall sentence score:}
Though $\text{DTW}_\text{SSL}$ exceeds human agreement, the most effective method for holistic scoring of Japanese sentences ensembles residual features and $\text{DTW}_\text{SSL}$, significantly exceeding human agreement ($r=66.0$ vs. $r=60.4$). Within this framework, $\text{DTW}_\text{norm. res}$ is the most predictive feature, accounting for 48.8\% of the explained variance in the Shapley decomposition.
The two distances are nearly collinear ($r=0.98$), but partial correlation analysis shows controlling for $\text{DTW}_\text{SSL}$, $\text{DTW}_\text{norm. res}$ retains a unique signal ($r_\text{partial}=0.19$) whereas controlling for $\text{DTW}_\text{norm. res}$ leaves $\text{DTW}_\text{SSL}$ with almost none ($r_\text{partial}=0.04$).
This may indicate that human raters evaluate overall pronunciation quality through a macro-perceptual lens, prioritizing structural prosody over localized phone accuracy.

\textbf{Intonation:}
The raters in this task are explicitly instructed to listen to pitch accent, but $F_0$ performs relatively poorly on its own.
Our best method for this task does not include $\text{DTW}_\text{norm. res}$ as with English; it uses a combination of $\sigma_\text{res}$ and $F_0$--intensity. Shapley analysis attributes 88.7\% of predictive power to $\sigma_\text{res}$ alone. However, the gap between the ensemble and $\sigma_\text{res}$ as a standalone feature is insignificant. This language difference suggests that the rated criteria in the JRF intonation task may differ substantially from those in the ERJ task (where $\sigma_\text{res}$ performs poorly), though the limited linguistic variety in this subset (only 8 texts) warrants caution.  

\textbf{Difficult sounds:}
As with English phonetic scoring of words, $\text{DTW}_\text{SSL}$ does not perform as well as the sentence counterpart. However, this task is particularly difficult, as human raters attend to specific parts of the words when rating, effectively ignoring the rest, whereas the DTW cost encapsulates the entire utterance.

\begin{table}[!t]
\renewcommand{\arraystretch}{1.0}
\setlength{\tabcolsep}{2.5pt}
\begin{threeparttable}
\caption{Evaluation results for tasks in the JRF~\cite{JRF} dataset (Japanese)}
\label{tbl:results-jrf}
\begin{tabularx}{\linewidth}{@{}X S[table-format=-2.1] S[table-format=-2.1] S[table-format=-2.1] S[table-format=1.2]@{}}
\toprule
\textbf{Task}/Method & {$r$ ($\uparrow$)} & {$\rho$ ($\uparrow$)} & {QWK ($\uparrow$)} & {MAE ($\downarrow$)}\\
\midrule
\textbf{Overall score (sentence)}\\
\hspace{2mm}\textit{Human benchmark} & 60.4 & 60.6 & 47.7 & 0.89 \\
\vspace{-0.08in}\\
\hspace{2mm}$\text{DTW}_\text{SSL}$ + $\text{DTW}_\text{norm. res}$ + $\sigma_{\text{res}}$$^{*\dagger}$ & \bfseries 66.0 & 65.6 & 50.8 & 0.68 \\
\hspace{2mm}$\text{DTW}_\text{SSL}$$^{\dagger}$ & 64.5 & 64.4 & 47.5 & 0.69 \\
\hspace{2mm}$\text{DTW}_\text{PPG}$ & 35.6 & 36.2 & 11.5 & 0.82 \\
\vspace{-0.07in}\\

\textbf{Intonation (sentence)}\\
\hspace{2mm}\textit{Human benchmark} & 47.4 & 48.5 & 38.5 & 0.87 \\
\vspace{-0.08in}\\
\hspace{2mm}$\sigma_{\text{res}}$ + $F_0$--intensity$^{*}$ & \bfseries 41.2 & 45.8 & 23.7 & 0.81 \\
\hspace{2mm}$\sigma_{\text{res}}$ & \bfseries 39.7 & 43.3 & 20.9 & 0.80 \\
\hspace{2mm}$F_0$ (baseline) & 17.3 & 29.7 & -0.0 & 0.97 \\
\hspace{2mm}$F_0$--intensity (baseline) & 13.8 & 16.7 & -0.2 & 0.96 \\
\vspace{-0.07in}\\

\textbf{Difficult sounds (word)}\\
\hspace{2mm}\textit{Human benchmark} & 47.8 & 43.8 & 32.3 & 1.35 \\
\vspace{-0.08in}\\
\hspace{2mm}$\text{DTW}_{\text{SSL}}$ & \bfseries 31.9 & 28.7 & 11.9 & 1.20 \\
\hspace{2mm}$\text{DTW}_\text{PPG}$ & 18.9 & 17.6 & 3.0 & 1.22 \\
\bottomrule
\end{tabularx}
\begin{tablenotes}[flushleft]
\footnotesize
\item $^{*}$Ensemble method. \textbf{Bold}: best or statistically equivalent (speaker-cluster bootstrap, $p>0.05$). $^{\dagger}$Significantly exceeds the human benchmark ($p<0.05$).
\end{tablenotes}
\end{threeparttable}
\end{table}

\section{Conclusion}

We showed that DTW over self-supervised representations proves a strong basis for multi-aspect pronunciation assessment.  It exceeds human inter-rater agreement at the sentence level for holistic and phonetic scoring. 
For rhythm, the warping path itself directly reflects the temporal deviations that rhythm perception is thought to rely on. 
To our knowledge, our proposed methods for rhythm are the current best for comparative rhythm assessment, approaching human performance.
Intonation remains the open challenge. Our methods provide a modest signal above acoustic baselines, but fall short of human agreement. Neither pitch-based features nor SSL prosodic residuals appear to capture what expert raters are actually responding to. 
Better disentanglement of prosodic information from SSL representations is the most pressing direction for future work, alongside evaluation across more languages.
\newpage

\section{AI-Generated Content Disclosure}
Generative AI was used for final proofreading of this manuscript. It was used for code generation, in particular dataset normalization, execution time optimization, and data visualization. All experiments were conceived by the authors and all final code was manually reviewed and edited for clarity.

\bibliographystyle{IEEEtran}
\bibliography{bibliography}

\end{document}